\documentclass{article}
\usepackage[preprint]{spconf}
\usepackage{amsmath,graphicx, cite, booktabs}
\usepackage{bm}
\usepackage{amsfonts, amssymb}
\usepackage{microtype}
\usepackage[subtle]{savetrees}

\copyrightnotice{\footnotesize \copyright~2021 IEEE}
\toappear{\footnotesize Published in the IEEE Spoken Language Technology Workshop (SLT), 2021.}

\usepackage{xcolor}
\definecolor{mycolor}{HTML}{FF9900}

\usepackage[prependcaption,textsize=scriptsize]{todonotes}
\usepackage{booktabs}
\usepackage{tabularx}
\usepackage{multirow}

\newcolumntype{C}{>{\centering\arraybackslash}X}
\newcolumntype{L}{>{\raggedright\arraybackslash}X}

\usepackage{cite}
\let\oldbibliography\thebibliography
\renewcommand{\thebibliography}[1]{\oldbibliography{#1}
	\setlength{\itemsep}{3pt}
	\vspace*{-0mm}}

\title{A comparison of self-supervised speech representations as input features for unsupervised acoustic word embeddings}

\name{Lisa van Staden \qquad Herman Kamper
\thanks{This work is supported in part by the National Research Foundation of South Africa (grant number: 120409) and a Google Faculty Award for HK.}}
\address{E\&E Engineering, Stellenbosch University \\
 		 {\small \texttt{18245471@sun.ac.za, kamperh@sun.ac.za}}}

\begin{document}
	\maketitle
	
\begin{abstract}
	Many speech processing tasks involve measuring the acoustic similarity between speech segments. 	\textit{Acoustic word embeddings} (AWE) allow for efficient comparisons by mapping speech segments of arbitrary duration to fixed-dimensional vectors. For zero-resource speech processing, where unlabelled speech is the only available resource, some of the best AWE approaches rely on weak top-down constraints in the form of automatically discovered word-like segments. Rather than learning embeddings at the \textit{segment level}, another line of zero-resource research has looked at representation learning at the \textit{short-time frame level}. Recent approaches include self-supervised predictive coding and correspondence autoencoder~(CAE) models. In this paper we consider whether these frame-level features are beneficial when used as inputs for training to an unsupervised AWE model. We compare frame-level features from contrastive predictive coding (CPC), autoregressive predictive coding and a CAE to conventional MFCCs. These are used as inputs to a recurrent CAE-based AWE model. In a word discrimination task on English and Xitsonga data, all three representation learning approaches outperform MFCCs, with CPC consistently showing the biggest improvement. In cross-lingual experiments we find that CPC features trained on English can also be transferred to Xitsonga.
\end{abstract}

	
\section{Introduction}

A number of speech processing tasks 
rely on measuring the acoustic similarity between speech segments~\cite{settle_query-by-example_2017,park_unsupervised_2008}.
Usually, similarity is measured using dynamic time warping (DTW), an algorithm that finds an
optimal alignment between speech segments~\cite{rabiner_considerations_1978}. 
However, DTW
is computationally expensive. 
This has {led} 
to research in methods of finding fixed-dimensional speech representations, referred to as
\textit{acoustic word embeddings} (AWEs)
\cite{levin_fixed-dimensional_2013, bengio_word_nodate, chung_audio_2016, kamper_deep_2016, kamper_truly_2019, maas_word-level_nodate,   chung_speech2vec:_2018, holzenberger_learning_2018}. 
These methods attempt to capture the acoustic information in speech segments of variable length and condense it in such a way that segments containing the same words are mapped to similar embeddings.
Since speech segments can then be represented in the same fixed-dimensional space, measuring the acoustic similarity can be done with a computationally inexpensive 
distance calculation.

Many of the downstream tasks for which AWEs are useful can be performed in a setting where transcribed speech resources are unavailable. 
Such tasks include query-by-example search~\cite{levin+etal_icassp15,huang+etal_arxiv18,yuan+etal_interspeech18}, where a speech segment is used as a query to search over a database of speech, and unsupervised term discovery (UTD)~\cite{park_unsupervised_2008,jansen_efficient_2011,ondel+etal_arxiv19,rasanen+blandon_arxiv20}, where the aim is to discover reoccurring patterns in untranscribed speech.
Speech processing in settings without any labelled speech data
is referred to as \textit{zero-resource speech processing} and it has become a popular field of research~\cite{jansen_summary_2013,versteegh_zero_2016, dunbar_zero_2017, dunbar_zero_2019}.
Apart from practical tasks, this area is also closely related to modelling language acquisition in humans, since infants acquire language without access to transcribed speech data~\cite{RASANEN2012975,elsner_speech_2017,matusevych+etal_cogsci20}.
A number of studies have specifi\-cally focussed on developing AWEs for this
zero-resource setting~\cite{levin_fixed-dimensional_2013, chung_audio_2016, kamper_truly_2019, van_staden_improving_2020, algayres_evaluating_2020}.
Several of these
unsupervised AWE approaches rely on weak top-down constraints in the form of discovered words from a UTD system.
For instance, the correspondence autoencoder recurrent neural network (CAE-RNN)~\cite{kamper_truly_2019} is trained to reconstruct one segment in a discovered pair given the other segment as input; embeddings are taken from an intermediate representation between the model's encoder and decoder RNNs.
By using discovered words as input-output pairs, the CAE-RNN can then be trained in the absence of any labelled speech data.

While AWE approaches attempt to model speech at the \textit{segment} level,
several zero-resource studies have focussed instead on unsupervised representation learning at the short-time \textit{frame} level~\cite{badino+etal_interspeech15,heck+etal_ieice18,tsuchiya+etal_icassp18,riad2018sampling,last+etal_spl20,vanniekerk+etal_interspeech20}.
Here the goal is to capture 
meaningful contrasts such as phone categories.
Many of these methods could be described as \textit{bottom-up}, learning representations directly from the lower-level features.
This includes recent self-supervised predictive coding methods~\cite{oord_representation_2018, chung_unsupervised_2019}.

In this paper we investigate and compare different learned speech representations 
as inputs to unsupervised AWE models.
For learning frame-level features, we consider three approaches. The first two are recent
predictive coding 
methods.
In contrastive predictive coding (CPC)~\cite{oord_representation_2018}, representations are learned by predicting the correct future frames from a set containing negative examples.
In autoregressive predictive coding (APC)~\cite{chung_unsupervised_2019},
representations are learned by predicting future frames with an autoregressive model.
Finally, we consider a frame-level correspondence autoencoder (CAE) neural network
\cite{kamper_unsupervised_2015}. 
We find pairs of speech segments that are predicted to be of the same type using a UTD system~\cite{jansen_efficient_2011}. 
DTW is then used to align frames between the pair of discovered speech segments.
The frame-level CAE model is then trained to predict corresponding aligned frames from each other.   

All three of these representation learning approaches attempt to capture linguistically meaningful 
information present at the short-time frame level 
(such as phone categories).
But they do so in very different ways. The CPC tries to discriminate between frames in an utterance from negative examples, while the APC tries to reconstruct future frames based on a past history.
The frame-level CAE follows quite a different methodology: trying to reconstruct frames from aligned segments predicted to contain the same word. 
As with the CPC and APC approaches, the CAE can also be described as self-supervised~\cite{algayres_evaluating_2020}, since labels for a proxy task are automatically obtained from the data~\cite{doersch+zisserman_iccv17}.

The three types of learned frame-level representations 
are used as input features to train unsupervised CAE-RNN AWE models.
We evaluate the intrinsic quality of the resulting AWEs in a word discrimination task.
We find that across two languages, English and Xitsonga, all three learned representations improve upon MFCCs, with CPC 
producing the best results when used as input features to the AWE model.
We also perform crosslingual experiments, where we find that using frame-level representation models trained on a higher resourced language (English) to encode those of a low-resource language (Xitsonga), improves the AWEs of the latter.

	\section{Self-Supervised Frame-Level Representation Learning}
\label{sec:framelevel}

\abovedisplayshortskip=1pt
\belowdisplayshortskip=1pt
\abovedisplayskip=6pt
\belowdisplayskip=6pt

\subsection{Contrastive predictive coding (CPC)}
\label{sec:cpc}

The aim of contrastive predictive coding (CPC) is to encode only the information that is shared between current and future {acoustic observations~\cite{oord_representation_2018}.
This results in representations that better describe shared short-time 
information, like phone categories or speech intonation, depending on how far ahead
future observations are.
The original CPC study showed
that it produced effective speech representation when
trained on raw audio waveforms \cite{oord_representation_2018}.
A more recent study \cite{blandon_analysis_nodate} showed that it can also be successfully applied when predicting conventional frame-level acoustic features.

Figure \ref{fig:cpc} shows the architecture for learning CPC representations.
A sequence of frames is received as input to the CPC model, with a single frame at time step $t$ denoted as $\bm{x}_t$.
The input frames are 
encoded by a function $g_{\text{enc}}$ into latent variables, denoted as 
$\bm{z}_t$ at time step $t$. 

\begin{figure}[t!] 
	\centering
	\includegraphics[width=0.95\linewidth]{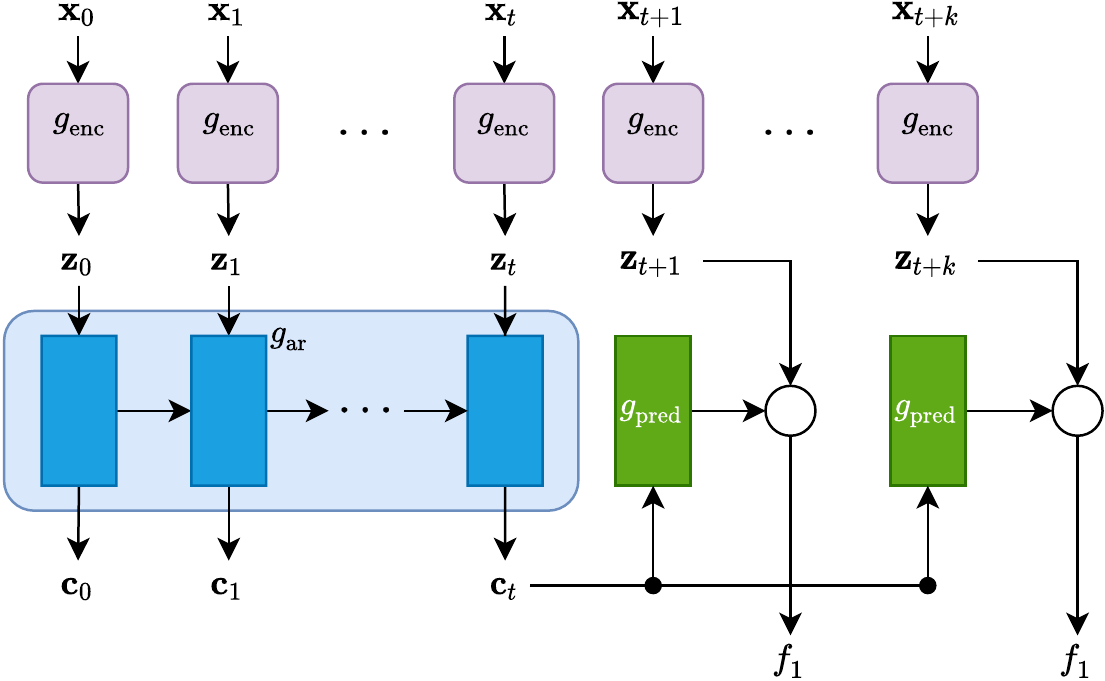}
	\caption{The CPC model is trained to compute a score from the context variables, $\bm{c}_0, \bm{c}_1, ... , \bm{c}_t$, and the future latent variables, $\bm{z}_{t+1}, ... , \bm{z}_{t+k}$.} 
	\label{fig:cpc}
\end{figure}

In our case this encoder function consists of a sequence of linear layers.
Next, the latent variables are encoded by an
autoregressive function $g_{\text{ar}}$ into context variables 
$\bm{c}_t$. 
We use a recurrent neural network (RNN) layer for this purpose.
Because this function is autoregressive it allows each $\bm{c}_t$ to be a summary of all $\bm{z}_{\le t}$, such that $\bm{c}_t = g_{\text{ar}}(\bm{z}_{\le t})$.

The next step is to determine a prediction score for every $\bm{c}_t$ at each prediction step. Let $K$ be the chosen number of steps that we want to predict into the future. Then 
for each step $k \in  [1, K]$ a function $g_{\text{pred}}^k$ is used to transform $\bm{c}_t$. 
A log bilinear function is then used to calculate the score:
\begin{equation}
	f_k(\bm{z}_t, \bm{c}_t) = \exp \left(
						 	  \bm{z}_{t+k}^{\top} g_{\text{pred}}^k(\bm{c}_t)
						 	  \right)
\end{equation}

Let $\mathcal{Z}_t$ be a set that contains the true $\bm{z}_t$ along with $N - 1$ negative samples of $\bm{z}_t$. 
We also calculate prediction scores for the negative samples. 
The model is 
then trained to maximise the score for $\bm{z}_t$ and minimise it for the negative samples. 
Concretely, the loss function
used to do this is based on noise-contrastive estimation (NCE)~\cite{gutmann_noise-contrastive_nodate}:
\begin{equation}
	L_{\text{InfoNCE}}(\bm{x}_t, k) = -\text{log}\left( 
	\frac{f_k(\bm{z}_t, \bm{c}_t)}{\sum_{\bm{z}_i \in Z_t} f_k(\bm{z}_i, \bm{c}_t)} 
	\right)
	\label{eg:contrastive_loss}
\end{equation}
The final loss function applied to the CPC model for a sequence $X$ of input frames can then be expressed as:
\begin{equation}
	L_{\text{CPC}} =  \frac{1}{K} \frac{1}{|X|}\sum_{k \in [1, K], \bm{x}_j \in X} L_{\text{InfoNCE}}(\bm{x}_j, k)
\end{equation}

We sample negative frames from
different utterances of the same speaker as the correct frames. 
This encourages the CPC model to normalise out speaker information, since it can't use this information to select the correct frame from among the negative examples~\cite{van_niekerk_vector-quantized_2020}.

Either} $\bm{z}_t$ or $\bm{c}_t$ can be used as frame-level representations for a downstream task. 
But it is recommended to use $\bm{c}_t$ when extra context from the past is useful \cite{oord_representation_2018}. 
In our development experiments 
$\bm{c}_t$ did give 
better results, and we therefore 
use it as our input representations to the AWE~models.

\subsection{Autoregressive predictive coding (APC)}
\label{sec:apc}

Similarly to CPC, the aim of autoregressive predictive coding (APC) is to encode only the information that is shared between current and future frames.
The original 
APC paper~\cite{chung_unsupervised_2019} argues that when learned representations are encouraged to throw out nuisance information (like speaker identity or noise) there is a risk that useful information might also be 
lost. 
So instead of encouraging the model to normalise
out non-discriminative features, as with the score maximisation of CPC, an autoregressive function is used to decode the predicted future frame from a latent variable containing more general information.
\begin{figure}[!t]
	\centering
	\includegraphics[width=0.39\textwidth]{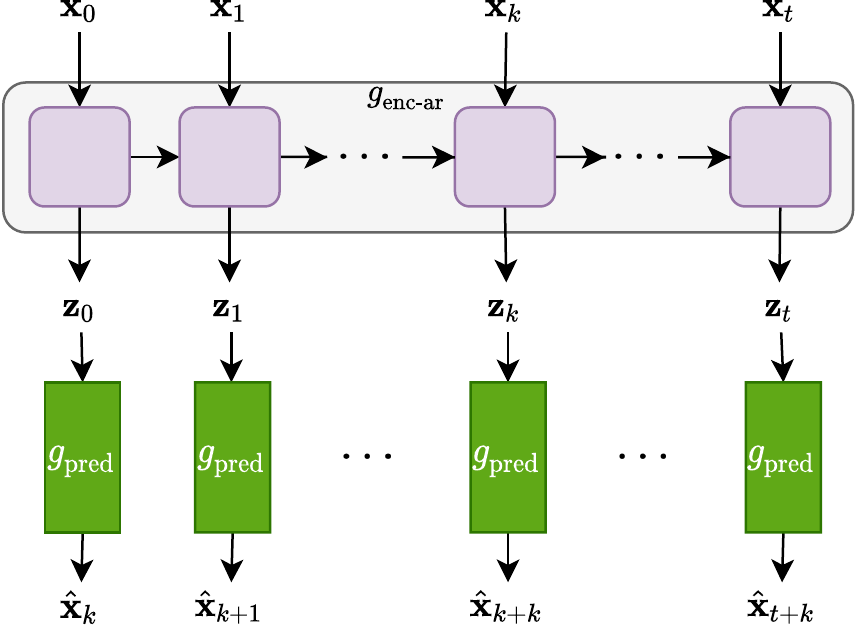}	
	\caption{The APC model is trained to predict
    the frame $k$ steps ahead of the input frame from a latent variable.}
	\label{fig:apc}
\end{figure}
Figure \ref{fig:apc} shows the APC architecture. 
A sequence of frames 
are encoded by an autoregressive function $g_{\text{enc-ar}}$.
In our case the autoregressive function is a stack of RNN layers. 
The last layer's hidden states at each time step is then used as the latent variables,
denoted as $\bm{z}_t$ for
time step $t$.
Next, a prediction function $g_{\text{pred}}$ transforms each $\bm{z}_t$ to the predicted input frame $k$ steps ahead, such that it can be described as  $\hat{\bm{x}}_{t + k} = g_{\text{pred}}(\bm{z}_t)$.
For a sequence of input frames $X = (\bm{x}_0, \bm{x}_1, \ldots, \bm{x}_T)$ the model is trained to minimise the mean absolute error (MAE) between the true and predicted future frames:
\begin{equation}
	L_\text{MAE}(X) = \frac{1}{|X|} \sum_{\bm{x}_t \in X} {|| \bm{x}_{t + k} - \hat{\bm{x}}_{t+k}||}_1
\end{equation}
A follow-up study on APC proposed adding an auxiliary loss as a regularisation term  \cite{chung_improved_2020}. 
This loss encourages the latent variables to include information from previous frames in the sequence.
Concretely,
$M$ different frames are chosen at random from $X$
to use as anchors. 
An anchor at position $m$ is denoted by $\bm{x}_{a_m}$. 
For each anchor we take a slice of $X$, denoted by $A_m$, that contains $n$ frames that start 
$s$ time steps behind $a_m$, such that
 $A_m = \left( \bm{x}_{a_m - s}, \bm{x}_{a_m - s + 1}, ..., \bm{x}_{a_m - s + n - 1} \right)$.
Then the auxiliary loss is given as
the MAE loss for every $A_m$: 
\begin{equation}
L_\text{aux}(X) = \frac{1}{M} \sum_{A_m \in \mathcal{A}} L_\text{MAE}(A_m)
\label{eq:apc_aux}
\end{equation}
where $\mathcal{A}$ denotes the set that contains every $A_m$ sequence sliced from $X$.

In our development experiments we found that adding the auxiliary loss does result in a small improvement for the AWEs. 
The final loss function for our APC model is therefore 
\begin{equation}
	L_{\text{APC}} = L_{\text{MAE}} + \lambda L_{\text{aux}}
\end{equation}
with $\lambda$ a hyper-parameter.

The hidden states from any of the layers of $g_{\text{enc-ar}}$ can be used as frame representations for downstream tasks. 
Previous research on autoregressive textual word embedding 
models showed that the information in hidden states are hierarchical across across layers~\cite{peters_dissecting_2018}.
The original APC study~\cite{chung_unsupervised_2019} concluded that earlier layers in the autoregressive function contain more speaker information and later layers more phonetic information. 
Therefore we use the hidden states of the last layer $\bm{z}_t$ as speech representations in our AWE experiments.

\subsection{Frame-level correspondence autoencoder (CAE)}
\label{sec:cae}

Unlike with the predictive coding models, the correspondence autoencoder (CAE) model has no autoregressive component~\cite{kamper_unsupervised_2015}. 
Therefore the model does not encode information that is shared temporally. 
The model is rather encouraged to encode only the information that is shared between frames from different instances of the same (predicted) word, as outlined 
detail
below.
The intuition is that this
encourages the model to normalise
out noise and speaker information, since these properties could be different for the input and output~frames.

\begin{figure}[!t]
    \centering
    \includegraphics[width=0.25\textwidth]{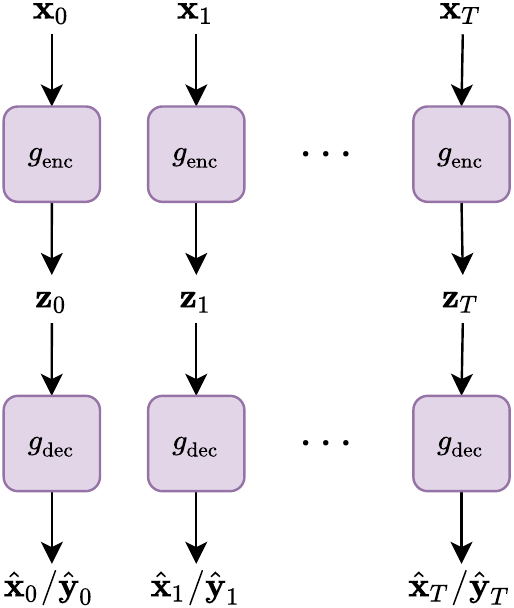}
    \caption{
    The AE is trained to reconstruct the input frames $\bm{x}_{0},\bm{x}_{1}, ... , \bm{x}_{T}$ directly.
    The CAE is trained to reconstruct frames from another segment $\bm{y}_{0},\bm{y}_{1}, ... , \bm{y}_{T}$, predicted to be a different instance of the same word as the input.
	}
    \label{fig:frame_cae}
\end{figure}

The CAE model produces the best results if its weights are initialised with those of a trained autoencoder (AE) \cite{kamper_unsupervised_2015, last+etal_spl20}. 
The architectures of these two models
are similar and are shown in Figure \ref{fig:frame_cae}.

The AE takes 
a frame $\bm{x}_t$ as input. 
A function $g_{\text{enc}}^{\text{(AE)}}$ encodes the input into a latent variable $\bm{z}_t$. 
This latent variable is then decoded by the function $g_{\text{dec}}^{\text{(AE)}}$, yielding the model output $\hat{\bm{x}}_t = g_{\text{dec}}^{\text{(AE)}}(\bm{z}_t)$. 
The target for this output is the input frame itself (from there the $\hat{\bm{x}}_t$ notation).
For a batch of input frames $X$ the AE is trained to minimise the mean square error (MSE) between the input and predicted frames.
\begin{equation}
L_{\text{AE}} = \frac{1}{|X|} \sum_{\bm{x}_t \in X} ||\bm{x}_t - \hat{\bm{x}}_{t}||^2_2
\end{equation}

For the CAE model, we first have to create pairs of similar frames. 
Given an unlabelled speech collection, 
we use a UTD system to find speech segments which are predicted to be of the same unknown type~\cite{park_unsupervised_2008}. 
Pairs of discovered word segments are then aligned using DTW, producing input-output frame pairs for the CAE. 
Since the segments may differ in length, some frames could be paired with multiple other frames.

The architecture of the CAE model is the same as that of the AE model, but instead of decoding $\bm{z}_t$ in order to predict the input frame itself, we use it to predict the corresponding frame in the pair.
Formally, for an input-output pair $(\bm{x}_t, \bm{y}_t)$, the model takes $\bm{x}_t$ as input, produces  the latent representation $\bm{z}_t = g_{\text{enc}}^{\text{(CAE)}}(\bm{z}_t)$, and then decodes $\bm{z}_t$ to obtain the model output $\hat{\bm{y}_t} = g_{\text{dec}}^{\text{(CAE)}}(\bm{z}_t)$.
For a batch of input-output frame pairs $Y$,
the model is trained to minimise the mean square error (MSE) between the input and output frames.
\begin{equation}
L_{\text{CAE}} = \frac{1}{|Y|} \sum_{(\bm{x}_t, \bm{y}_t) \in Y} ||\bm{y}_t - \hat{\bm{y}}_{t}||^2_{2}
\end{equation}
We use the latent variables $\bm{z}_t$ from the CAE model as the representations in our AWEs.

\section{Acoustic Word Embedding Model}
\label{sec:awe}

Above we introduced three frame-level representation learning methods.
We will use each of these methods to produce input features for an unsupervised AWE model, and then compare the results.
The idea is that the resulting AWE model would be better able to discriminate at the segment (word) level by taking advantage of features learned at the 
frame level.

Concretely, we will use the correspondence autoencoder recurrent neural network (CAE-RNN) AWE model of~\cite{kamper_truly_2019}.
Note that, although they are related, this AWE model is different from the frame-level CAE of Section~\ref{sec:cae}.
The unsupervised CAE-RNN was shown to give comparable or slightly better performance compared to a DTW approach~\cite{kamper_truly_2019}, making it one of the best unsupervised AWE models.

\begin{figure}[!t]
	\centering
	\includegraphics[width=0.3\textwidth]{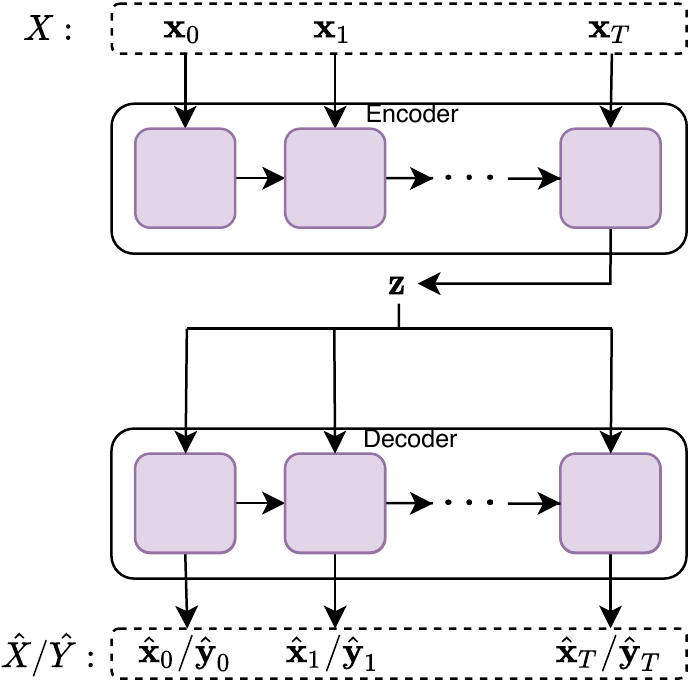}
	\caption{
	The AE-RNN is trained to reconstruct
	 input sequence $X = (\bm{x}_{0},\bm{x}_{1}, ... , \bm{x}_{T})$ from latent variable $\bm{z}$.
	The CAE-RNN
	is trained to reconstruct a different sequence $Y = (\bm{y}_{0},\bm{y}_{1}, ... , \bm{y}_{T})$ predicted to contain the same word as input $X$.
	}
	\label{fig:cae_rnn}
\end{figure}

The weights of the CAE-RNN are initialised with those of a autoencoder recurrent neural network (AE-RNN) \cite{chung_audio_2016}.
Both models are based on encoder-decoder model architecture \cite{sutskever_sequence_2014, cho_learning_2014}, as illustrated in Figure \ref{fig:cae_rnn}.
The encoder and decoder each consists of a stack of RNN layers. 
The encoder maps an input sequence $X$ 
of variable length into a fixed-dimensional latent variable $\bm{z}$. 
This latent variable could be the last hidden state
of the last encoder RNN layer, but in our case we add a linear layer after the encoder to transform the last hidden state into $\bm{z}$.
We use these latent variables $\bm{z}$ as acoustic embeddings.
The decoder then maps $\bm{z}$ to an output sequence, denoted by $\hat{X}$ for the AE-RNN and $\hat{Y}$ for the CAE-RNN.

The AE-RNN is trained so that $\hat{X}$ gives a reconstruction of the original
input sequence~\cite{chung_audio_2016}.
We do this by minimising the MSE between the true and reconstructed sequences:
\begin{equation}
L_{\textrm{AE-RNN}} = \frac{1}{|X|} ||X-\hat{X}||^2_{2,1}
\label{eq:AE-RNN}
\end{equation}

Instead of reconstructing the input sequence, the CAE-RNN is trained to reconstruct another instance of the same (predicted) word as the input sequence~\cite{kamper_truly_2019}.
Since our training data is unlabelled, we again (like with the frame-level CAE in Section \ref{sec:cae}), use a UTD 
system to automatically discover speech segments which are predicted to be of the same type.
For a given pair of sequences $(X, Y)$, the CAE-RNN is fed with input $X$ and then trained to reconstruct $Y$ at its output.
We do this by minimising the MSE between the true sequence $Y$ and the predicted output $\hat{Y}$.
\begin{equation}
L_{\textrm{CAE-RNN}} = \frac{1}{|Y|} ||Y - \hat{Y}||^2_{2,1}
\label{eq:CAE-RNN}
\end{equation}
The intuition behind the CAE-RNN is that the model learns to only encode information that is shared between the input-output segments (such as the word identity) while throwing out nuisance 
information. {This is similar to the idea behind the frame-level CAE (Section \ref{sec:cae}) but here we use whole segments containing a sequence of frames instead of single frames.}

    \section{Experimental Setup and Evaluation}

\subsection{Data}

As in~\cite{kamper_truly_2019}, we
train our models on datasets from two languages: English, from the Buckeye corpus \cite{pitt_buckeye_2005}, and Xitsonga, from the NCHLT corpus \cite{vries_smartphone-based_2014}.
The English training, validation and test sets each contain around six hours of speech.
For Xitsonga, we have a single set of 2.5~hours.
For both the frame-level CAE (Section~\ref{sec:cae}) and segment-level CAE-RNN (Section~\ref{sec:awe}), we
use the UTD system of~\cite{jansen_efficient_2011} to discover pairs of speech segments.
In the English training set, 14k unique pairs are discovered.
In the Xitsonga
set, around 6k unique pairs are discovered. For the CPC model we assume that speaker labels are available.
As in~\cite{versteegh_zero_2016,kamper_truly_2019}, no validation data is available for Xitsonga; we therefore perform all development experiments on the English validation data and then use exactly the same hyperparameters on the Xitsonga set, replicating a true zero-resource setting.

All speech audio are transformed to 13-dimensional Mel-frequency cepstral coefficients (MFCCs).
These are used as input to the predictive coding feature learning models (Sections~\ref{sec:cpc} and~\ref{sec:apc}).
For the frame-level CAE (Section~\ref{sec:cae}), it was found beneficial to additionally include velocity and acceleration coefficients (39-dimensional input). 

\subsection{Representation learning model implementations}

\begin{table*}[!t]
	\vspace*{-7pt}
	\caption{AP (\%) results on the English and Xitsonga test data for DTW and the CAE-RNN and downsampling AWE approaches. On the right is the crosslingual AP (\%) results obtained by training the frame-level representations on English before applying it to the Xitsonga data to train a CAE-RNN AWE model.}
	\vspace{5pt}
	\renewcommand{\arraystretch}{1.1}
	\begin{tabularx}{1\linewidth}{lCCC|CCC|c}
		\toprule
		& \multicolumn{3}{c}{English} & \multicolumn{3}{c}{Xistonga} & Crosslingual \\
		\cmidrule(lr){2-4} \cmidrule(lr){5-7} \cmidrule(l){8-8}
		& DTW & Downsampled & CAE-RNN & DTW & Downsampled & CAE-RNN & CAE-RNN \\
		\midrule
		MFCC & $ 35.90 $ & $ 19.40 $ & $30.18 \pm 0.34 $ & $ 28.15 $ & $ 18.36 $ & $ 22.52 \pm 0.29 $ & -- \\
		CPC & $16.03 \pm 0.03 $ &  $ 25.38 \pm 0.48 $ & $ \mathbf{36.83 \pm 0.92 } $ & $ 7.65 \pm 0.32 $ & $18.66 \pm 1.24$ & $ \mathbf{40.93 \pm 0.77} $ & $ \bm{41.79} \pm \bm{0.60} $ \\
		APC & $ 30.68 \pm 0.26 $ & $ 20.48 \pm 0.08 $ & $ 33.55 \pm 1.03 $ & $ 18.65 \pm 0.42 $ & $ 16.16 \pm 0.22 $ & $ 38.96 \pm 0.74 $ & $ 40.07 \pm 1.13 $ \\
		CAE & $ 41.49 \pm 1.97 $ & $ 21.27 \pm 1.51 $ & $ 31.31 \pm 1.17 $ & $ 48.32 \pm 1.96 $ & $ 26.01 \pm 0.33 $ & $ 29.61 \pm 3.21 $ & $ 34.25 \pm 1.61 $ \\
		\bottomrule
	\end{tabularx}
	\label{tab:english}
	\vspace*{-5pt}
\end{table*}

For our CPC model (Section~\ref{sec:cpc}), we use an encoder
of six 512-unit linear layers with layer normalisation and ReLU activation functions in between. 
A dropout layer with a rate of 0.5 is added after the third ReLU activation. 
In development experiments we found that the dropout layer does not improve results, but it does stabilise 
training.
We choose a long-short term memory (LSTM) layer as our summarising autoregressive function \cite{hochreiter_long_1997}. The dimensions of $\bm{z}_t$ and $\bm{c}_t$ are 64 and 356, respectively.
For the contrastive loss in~\eqref{eg:contrastive_loss}, based on validation experiments, we choose 31 negative examples from the same batch and we predict three steps ahead.
The model is trained with a learning rate of $1 \cdot 10^{-5}$. 
All our neural networks are optimised using Adam~\cite{kingma_adam:_2014}.
Each batch contains nine utterances from nine different speakers. 
We train the English model for a maximum of 15k
epochs, but stop at the epoch that produced the best 
results on validation data. 
We find that this happens when the model is at a training loss of around 0.93, and so we train the Xitsonga model until it reaches
this loss value.

The autoregressive encoder of our APC model (Section~\ref{sec:apc})
consists of a stack of three gated recurrent unit (GRU) \cite{cho_learning_2014} layers with a hidden state dimensionality of 512, which is thus also the dimensionality of $\bm{z}_t$. 
The predictor function is one linear layer which, based on validation experiments, 
predicts two steps ahead.
For the auxiliary loss~\eqref{eq:apc_aux}, we choose twelve anchors that we use to create sequences of seven frames from 14 steps back and we predict five steps ahead. 
We use an auxiliary loss  weight of $\lambda=0.1$.
In development experiments we found the number of epochs that produces the best AWEs on
the English validation data to be 50, and also use this many epochs on the Xitsonga data. 
We use a learning rate of $1 \cdot 10^{-3}$.

We set up our frame-level AE and CAE models (Section~\ref{sec:cae}) as in \cite{last+etal_spl20}.
The encoder and decoder functions both consist of six 100-unit linear layers with a 39-dimensional latent variable in between.
Through development experimentation we found that on the English datasets the best results are achieved if the AE model is trained for
five epochs and the CAE for ten epochs, and therefore again do the same on Xitsonga.
A learning rate of $1 \cdot 10^{-3}$ is used for both models.
 
\subsection{Unsupervised AWE model implementation}

We compare
MFCCs, CPC, APC and CAE representations (Section~\ref{sec:framelevel}) as input features to 
the unsupervised CAE-RNN AWE model (Section~\ref{sec:awe}).
This model is pre-trained as an AE-RNN using~\eqref{eq:AE-RNN} before switching to the CAE-RNN loss of~\eqref{eq:CAE-RNN}.
We follow the model setup of \cite{kamper_truly_2019}.
The dimensionality of the embeddings is set to 130. 
The encoder and decoder functions each consist of a stack of three GRUs with a hidden state dimensionality
of 512. 
We use learning rates of $1 \cdot 10^{-3}$ and $1 \cdot 10^{-4}$ for the AE-RNN and CAE-RNN, respectively, and both models use a batch size of 256. 
For the English dataset, we train the AE-RNN for 150 epochs and the CAE-RNN for 25 epochs and use early-stopping on validation data. 
For the Xitsonga dataset, we do not have validation data, so we average the number of epochs that it takes to produce the best AWEs on the English validation data for each of the different types of input representations.

\subsection{Evaluation}

We use the same-different task to evaluate the intrinsic quality of the AWE models \cite{carlin_rapid_nodate}. 
This task works as follows.
First an evaluation set of isolated words is embedded using a particular AWE model.
A decision of whether two embedded segments contain the same or different words can then be made based on the distance between the embeddings. By varying a threshold, we
create a curve of precision versus recall measuring correct predictions. 
The average precision~(AP) 
is the area under this curve. 
We use AP
to measure the quality of the AWEs, where higher scores are better.

As an AWE baseline, we use downsampling~\cite{levin_fixed-dimensional_2013,holzenberger_learning_2018}.
In this method, we choose
ten equally spaced frame representations from a sequence and interpolate them to form an AWE.
We obtain downsampled AWEs from each of the representation learning methods considered.
As an additional way of performing the same-different task, we also consider using DTW over the  speech segments, each segment represented using the frame-level representation under consideration.
This approach 
has access to the full sequences without any compression.

All experiments are repeated three times.
We report
averaged AP scores along with the standard deviations.

	\section{Experiments}

Our main research question is whether improved self-supervised frame-level feature learning is beneficial when used in combination with a segment-level model for producing AWEs.
We therefore compare MFCC, CPC, APC and CAE features when used as input representations to an CAE-RNN model.
For comparison, we also use these frame-level features in downsampled AWEs as well as a direct DTW approach.
Additionally, 
we are 
interested to see if the learned representations can be used across languages.
This is related to previous studies applying frame-level features learned on one language to another, e.g.~\cite{hermann+goldwater_interspeech18,conneau_unsupervised_2020,riviere_unsupervised_2020}.
But here we are specifically interested in the 
resulting AWEs, which have not been considered before. We train the frame-level representations on English data and then use the trained models to encode the Xitsonga data is then fed to the CAE-RNN. 

Table \ref{tab:english} shows the AP scores of the English, Xistonga and crosslingual test experiments.
First focusing only on the AWE results of the English and Xitsonga experiments (downsampling and CAE-RNN columns), we see that in both cases  all the
learned representations improve upon MFCCs.
In both AWE approaches, best results are consistently achieved when using
the CPC representations.
Overall, the best AWE approach is the CAE-RNN taking in CPC representations as input, outperforming the downsampled CPC AWEs by more than 10\% and 20\% absolute in~AP for the English and Xitsonga data, respectively.

Somewhat surprisingly, when the features are used directly to do the same-different task (DTW columns), the only features to outperform MFCCs, for both languages, are the frame-level CAE features.
Moreover, for both the CPC and APC features, the corresponding CAE-RNN actually outperforms its DTW counterpart (e.g.\ 16.03\% vs.\ 36.83\% for the English CPC features and 38.96\% vs.\ 18.65\% for the Xitsonga APC features).
This is despite DTW having access to the full sequences while the CAE-RNN needs to compress the sequences into an AWE.
One reason for this could be that the weak top-down constraints used in the frame-level CAE are the same as those used in the CAE-RNN model (obtained from the UTD system), and therefore does not provide any additional signal.
In contrast, the self-supervision signal for the CPC and APC models are obtained in a bottom-up fashion which is different from that of the CAE-RNN---the top-down signal used in the CAE-RNN seems to be complementary to the bottom-up approach of CPC and APC.
But these are speculations and further investigation is required.

The right-most column shows the cross-lingual test results of the resulting AWEs. Again, the CPC features perform best. Surprisingly, the representations learned on English perform better than using those trained on Xitsonga.
The reason for this is 
likely due to the English dataset containing more 
speech data.
There is therefore potential for even larger improvements by using more substantial amounts of unlabelled data, as e.g.\ in~\cite{baevski2020wav2vec}.

\begin{table} 	
	\vspace*{-7pt}
	\renewcommand{\arraystretch}{1.1}
	\caption{Speaker classification accuracy (\%) results of the English development AWEs}
	\centering
	\vspace{5pt}
	\begin{tabular}{c c}
		
		\toprule
		& Speaker acc. (\%) \\ 
		\midrule
		MFCC & $ 64.05 \pm 3.89 $ \\
		CPC & $ 57.83 \pm 1.65 $ \\
		APC & $ 60.76 \pm 3.20 $  \\
		CAE & $ 51.43 \pm 1.95 $ \\
		
		\bottomrule
	\end{tabular}
	\label{tab:speaker}
	\vspace*{-7pt}
\end{table}

Finally, we consider a speaker information probing task on the English development data. 
A linear classifier is used to predict the speaker of the final AWEs produced by the CAE-RNN for each type of frame-level feature.
The speaker classification results, seen in Table \ref{tab:speaker}, show that all the learned representations lead to reduced speaker information in the AWEs compared to those produced by the MFCCs, with the CAE representations leading to the biggest reduction.

	\section{Conclusion}

We considered how unsupervised acoustic word embedding (AWE) models can be improved by taking in frame-level features from self-supervised approaches.
Concretely, we compared contrastive predictive coding (CPC), autoregressive coding and a correspondence autoencoder (CAE) when used as input to a recurrent CAE-based AWE model.
In a word discrimination task on two languages 
CPC features outperformed MFCCs as well as the other learned representations.

However, different trends were observed when using the features to perform the task directly using DTW: in this case, CPC performed worst while the frame-level CAE performed best.
Future work will investigate this discrepancy.
This might also be related to recent work~\cite{algayres_evaluating_2020} showing the limitations of the same-different task as an intrinsic AWE metric.
Future work will therefore also consider other downstream tasks.

	\bibliographystyle{bib/IEEEbib}
	\bibliography{bib/ref}
\end{document}